\documentclass{article}
\usepackage{spconf,amsmath,graphicx,hyperref}
\usepackage{booktabs}
\usepackage[table]{xcolor}
\usepackage{booktabs}
\usepackage{multirow}
\usepackage{makecell}
\usepackage{graphicx}
\usepackage{amsmath}
\usepackage{graphicx}
\usepackage{booktabs}
\usepackage{multirow}
\usepackage{tabularx}
\usepackage{xcolor}
\usepackage{colortbl}
\usepackage{url}
\usepackage{wrapfig}
\usepackage{caption} 
\usepackage{enumitem}

\title{Complementary Roles of Activation and Parametric Memory in Few-Shot Learning}
\name{%
\begin{tabular}{@{}c@{}}
Miaohe Niu$^{1,*}$,
Runsong Zhao$^{1,*}$,
Xinyu Liu$^{1}$,
Bo Jin$^{1}$,\\
Yucheng Qiao$^{1}$,
Chunliang Zhang$^{1,2}$,
Jingbo Zhu$^{1,2}$,
Tong Xiao$^{1,2,\dagger}$,
\end{tabular}%
\thanks{%
\makebox[1em][l]{$*$}Equal contribution.\protect\\
\hspace*{1.8em}\makebox[1em][l]{$\dagger$}Corresponding author.}
}

\address{%
$^{1}$ Northeastern University, Shenyang, China\\
$^{2}$ NiuTrans Research, Shenyang, China
}

\begin{document}
\ninept
\maketitle
\begin{abstract}

At test time, large language models (LLMs) can encode historical information in activation memory (i.e., KV caches) and parametric memory (i.e., updated parameters). While activation memory is generally considered effective for factual recall and parametric memory for learning new tasks, their interplay remains unclear. In this work, we systematically investigate the role of memory in few-shot learning through controlled experiments. We find that activation memory is superior for recalling facts, whereas parametric memory does not consistently outperform activation memory in task learning. 
Moreover, our experiments show that the composite task, Conditional Arithmetic, requires the synergy of both memory types. 
Through neuron-level analysis, we find that the model activates distinct sets of neurons when accessing the same historical information through activation versus parametric memory. 
When both memory types are combined, the model recruits neurons from both sets, which is crucial for solving Conditional Arithmetic. These findings suggest that neither memory mechanism alone is sufficient for this composite task, highlighting the importance of their collaboration.

\end{abstract}
\section{Introduction}
\label{sec:intro}

Memory is fundamental to the ability of large language models (LLMs) to acquire knowledge, learn from experience, and reason in new situations~\cite{geva-etal-2021-transformer}.
Understanding the mechanisms through which past information becomes available for subsequent learning and reasoning is therefore essential to developing more adaptable LLMs.
At test time, we distinguish two memory substrates for the purposes of this study: activation memory, which encodes history in contextual activations such as key-value (KV) caches, and parametric memory, which incorporates history through parameter updates via test-time training (TTT)~\cite{sun2020testtime,akyurek2025surprising}.
Prior work motivates a natural hypothesis: activation memory is better suited to factual recall, whereas parametric memory is better suited to learning new tasks~\cite{zheng2023editfactualknowledgeincontext,gozeten2026testtimetrainingprovablyimproves}.

However, whether this division of roles consistently holds remains unclear~\cite{pan-etal-2023-context,ghosh-etal-2026-fine}, as does how the two memory types work together.
To address these open questions, we investigate three research questions:
\textbf{RQ1:} \textit{Is activation memory actually better at factual recall, while parametric memory is better at learning new tasks?}
\textbf{RQ2:} \textit{Are there scenarios where effective task performance requires collaboration between the two memory types?}
\textbf{RQ3:} \textit{If so, what mechanisms underlie this collaboration?}

\begin{figure}[!t] 
    \centering
    \includegraphics[width=0.97\columnwidth]{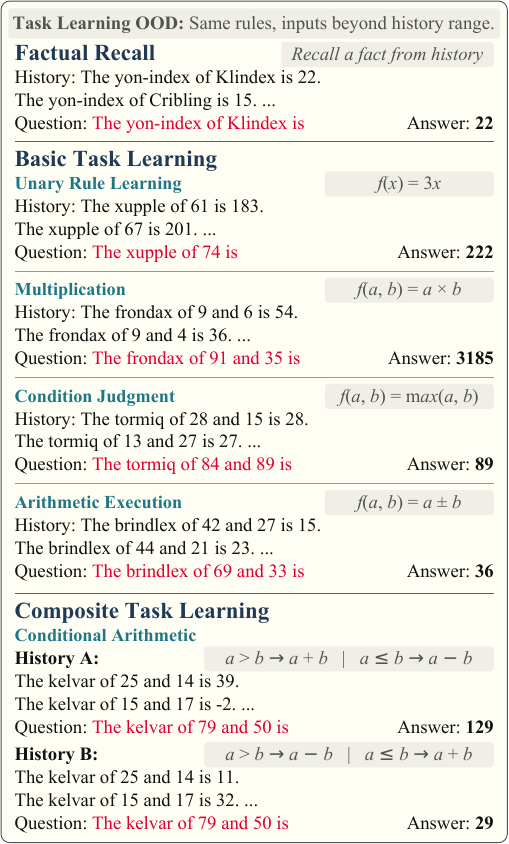}
    \caption{Examples of our synthetic tasks, grouped into factual recall, basic task learning, and composite task learning. Task-specific meanings and rules are defined by the supplied history rather than pretrained knowledge.}
    \label{fig:task_examples}
    \vspace{-7pt}
\end{figure}

To address these questions, we need to isolate information acquired at test time from knowledge already encoded in the base model. 
We therefore construct synthetic tasks involving novel concepts represented by invented words, whose task-specific meanings are available only through test-time history~\cite{pan-etal-2023-context,garg2022incontext}. 
History examples define each invented word through either associated facts or input--output demonstrations. 
For example, to compute the \textit{brindlex} of 42 and 27, the model must first infer from history examples what operation \textit{brindlex} represents (Fig.~\ref{fig:task_examples}). 
We compare activation-only, parameter-only, and combined memory settings using the same base model and histories.
For \textbf{RQ1}, our experiments confirm the advantage of activation memory in factual recall, but show that parametric memory is not consistently superior in learning new tasks.
For \textbf{RQ2}, we identify a conditional arithmetic task (see the last task in Fig.~\ref{fig:task_examples}) requiring the model to compare two numbers and then choose addition or subtraction according to a learned rule, where combining the
memories substantially outperforms either alone.

For \textbf{RQ3}, neuron-level analysis~\cite{dai-etal-2022-knowledge,wang-etal-2022-finding-skill} suggests how the two memory types collaborate.
When history is encoded only in parameters, neurons associated with condition judgment show weaker activity than in the combined setting.
When history is retained only in activations, neurons associated with arithmetic execution show weaker activity instead. 
Combining the memories yields stronger activity in both groups alongside improved task performance.
These results suggest that the two memory types support different steps of the composite task despite receiving the same history, indicating their complementary roles in task execution.

\section{Preliminaries}
\label{sec:preliminaries}

Surveys of memory in LLMs commonly distinguish parametric memory---knowledge held in model weights---from memory carried by the model's contextual state, although the latter is named differently across works~\cite{du2025rethinking,zhang2024memorysurvey,zhang2025memorymechanisms,xu2024knowledgeconflicts}.
When historical text is supplied as context, the model encodes it into activations used for inference.
Therefore, we refer to these two carriers as \emph{activation memory} and \emph{parametric memory}, respectively, following the distinction between working memory in activations and knowledge in weights~\cite{li-etal-2023-large}.

\subsection{Activation Memory}
Activation memory retains information in the states produced when a
model reads history. In autoregressive Transformers~\cite{vaswani2017attention},
these states are the key and value vectors cached for subsequent tokens;
in RNNs, they are recurrent hidden states~\cite{elman1990finding}.
Thus, contextual information can remain available to answer a new
question without updating the model parameters.
Activation memory is commonly expected to retain specific facts and
examples supplied in the current context~\cite{li-etal-2023-large}.

\subsection{Parametric Memory}
Parametric memory incorporates history through parameter updates.
One common use is inference-time adaptation: the model uses the current history to update its parameters or a temporary parameter module before answering the query~\cite{sun2020testtime,akyurek2025surprising,wang2024templora}.
Our parameter-only and combined settings use this form of adaptation.
Parametric memory is commonly expected to encode reusable rules that transfer across inputs~\cite{akyurek2025surprising,pan-etal-2023-context}.

A second line of work trains the model itself to perform such updates at test time. 
In TTT, for example, history updates the weights of a small internal learner. 
Those weights carry the sequence information as a state, while the surrounding model parameters are trained to determine how the state is initialized and updated~\cite{sun2025ttt}. 
DeltaNet and Titans use the same broad idea: history is written into a compact, learnable state rather than remaining only in the activations of a fixed model~\cite{yang2024delta,behrouz2025titans}. 
These approaches make test-time learning part of the model's training design, whereas we use inference-time adaptation to control whether history is encoded in activations, parameters, or both.

\subsection{Memory Encoding and Inference}
\label{sec:memory_settings}
Let $\mathcal{H}$ denote the history of $n$ examples, $q$ the query,
$a$ its target answer, and $\theta_0$ the pretrained model parameters.
Write $m_{\mathcal{H}}^\theta$ for the activation memory obtained by
encoding history $\mathcal{H}$ with parameters $\theta$.
Parameter adaptation instead gives
$\theta_{\mathcal{H}}=\mathcal{U}(\theta_0,\mathcal{H})$, where
$\mathcal{U}$ denotes the update procedure.

\begin{figure*}[t]
    \centering
    \includegraphics[width=\textwidth]{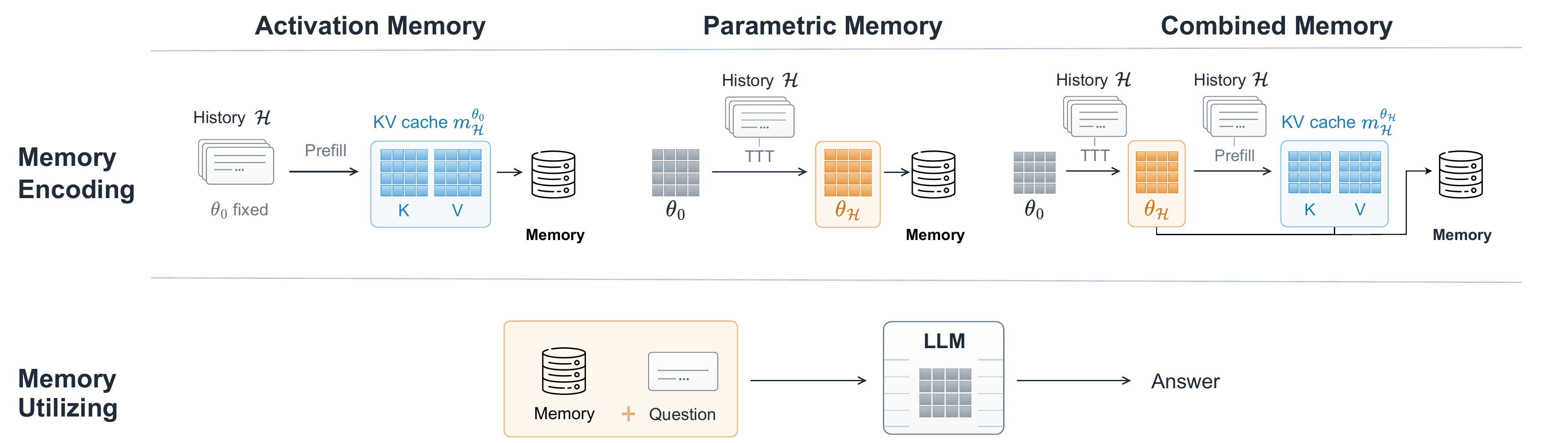}
    \caption{The three memory settings. \textsc{Act} encodes history in activations with fixed parameters; \textsc{Para} encodes it in adapted parameters without retaining history activations; \textsc{Combo} adapts the parameters and then re-encodes the same history into activations.}
    \label{fig:memory_routes}
\end{figure*}

Fig.~\ref{fig:memory_routes} illustrates three experimental settings,
each represented by a complete memory state $M$ containing the model
parameters and any retained history activations.

\textsc{Act} uses $M_{\mathrm{act}}=(\theta_0,m_{\mathcal{H}}^{\theta_0})$:
history $\mathcal{H}$ is retained in activations with the base parameters
fixed.

\textsc{Para} uses $M_{\mathrm{para}}=(\theta_{\mathcal{H}},\emptyset)$:
history is written into the parameters. Here $\emptyset$ means no
history-derived activation state is retained; parametric memory is
still present.

\textsc{Combo} uses
$M_{\mathrm{combo}}=(\theta_{\mathcal{H}},m_{\mathcal{H}}^{\theta_{\mathcal{H}}})$:
after adaptation, the same history is encoded again with the adapted
model. Its activation memory therefore need not equal that of \textsc{Act}.

All settings generate an answer as
$\hat a=\operatorname{LLM}(q;M)$, where $\operatorname{LLM}$ denotes
autoregressive generation using the parameters and any history state
contained in $M$. We use the same query and decoding procedure;
each instance starts from $\theta_0$.

\subsection{Task Definitions}
\label{sec:task_definitions}
We define six tasks for factual recall, rule learning, and their composition. 
Invented words and rules specified only through test-time examples keep the required instance-specific information out of pretraining, so the evaluated facts and mappings must be obtained from the supplied history. 
Across settings, the history, query, and decoding procedure are fixed.

Fig.~\ref{fig:task_examples} shows six tasks in three groups.
\emph{Factual Recall} asks for a fact explicitly supplied in history.
\emph{Basic Task Learning} requires applying a demonstrated rule: Unary Rule Learning maps one number to another; Multiplication computes a product; Condition Judgment selects an input by comparison; and Arithmetic Execution applies addition or subtraction.
Unlike \emph{Factual Recall}, which isolates retrieval, the four basic tasks each require a single learned operation, testing the RQ1 hypothesis that the two memory types have different strengths~\cite{garg2022incontext}.
\emph{Composite Task Learning} uses Conditional Arithmetic: the model judges a condition, selects addition or subtraction, and executes it in one instance. 
This tests whether either memory type suffices or both are needed, as posed by RQ2.

\section{Experiments}
\label{sec:experiments}


\subsection{Experimental Setup}
\label{sec:experimental_setup}

We evaluate Qwen3-1.7B-Base, Qwen3-4B-Base, and
Qwen3-8B-Base~\cite{qwen2025qwen3}, together with
Llama-3.2-1B~\cite{meta2024llama32}, on the six
synthetic tasks. The main experiment uses 300
evaluation instances per task, each with $n=50$
history examples.
We report exact-match accuracy against the target answer $a$.

To encode history in parameters, we update all model parameters using SGD with Nesterov momentum of 0.9, a learning rate of $10^{-2}$, and a gradient clipping threshold of 1.0. 
We optimize a causal language-modeling loss over $\mathcal{H}$ for 2 or 20 steps. 
The 2-step setting follows the two-epoch adaptation of Temp-LoRA~\cite{wang2024templora}, with each step using the full history, while 20 steps test whether the results persist under more extensive adaptation.

\subsection{Results for RQ1}
\label{sec:cross_task_results}

\begin{table*}[t]
\centering
\begingroup

\fontsize{9}{10.2}\selectfont
\setlength{\tabcolsep}{3.2pt}
\renewcommand{\arraystretch}{1.08}

\begin{tabular*}{\textwidth}{@{\extracolsep{\fill}}lcccccc@{}}
\toprule

& & \multicolumn{4}{c}{\textbf{Basic Task Learning}}
& \makecell{\textbf{Composite Task}\\\textbf{Learning}} \\

\cmidrule(lr){3-6}
\cmidrule(lr){7-7}

\textbf{Memory}
& \makecell{\textbf{Factual}\\\textbf{Recall}\\(2/20)}
& \makecell{Unary Rule\\Learning\\(2/20)}
& \makecell{Multiplication\\(2/20)}
& \makecell{Condition\\Judgment\\(2/20)}
& \makecell{Arithmetic\\Execution\\(2/20)}
& \makecell{Conditional\\Arithmetic\\(2/20)} \\

\midrule
\rowcolor{gray!18}
\multicolumn{7}{c}{\textit{Qwen3-1.7B-Base}} \\
\midrule

\textsc{Act}
& \mbox{\textbf{1.000}\,/\,\textbf{1.000}}
& \mbox{\textbf{1.000}\,/\,\textbf{1.000}}
& \mbox{\textbf{0.527}\,/\,\textbf{0.513}}
& \mbox{0.463\,/\,0.473}
& \mbox{0.770\,/\,0.790}
& \mbox{0.257\,/\,0.257} \\

\textsc{Para}
& \mbox{0.000\,/\,0.020}
& \mbox{0.000\,/\,0.800}
& \mbox{0.113\,/\,0.310}
& \mbox{0.003\,/\,0.200}
& \mbox{0.947\,/\,0.733}
& \mbox{0.413\,/\,0.390} \\

\textsc{Combo}
& \mbox{0.973\,/\,0.920}
& \mbox{\textbf{1.000}\,/\,0.927}
& \mbox{0.467\,/\,0.213}
& \mbox{\textbf{0.613}\,/\,\textbf{0.650}}
& \mbox{\textbf{1.000}\,/\,\textbf{0.927}}
& \mbox{\textbf{0.637}\,/\,\textbf{0.877}} \\

\midrule
\rowcolor{gray!18}
\multicolumn{7}{c}{\textit{Qwen3-4B-Base}} \\
\midrule

\textsc{Act}
& \mbox{\textbf{1.000}\,/\,\textbf{1.000}}
& \mbox{\textbf{1.000}\,/\,\textbf{1.000}}
& \mbox{0.657\,/\,0.663}
& \mbox{0.733\,/\,\textbf{0.737}}
& \mbox{0.740\,/\,0.727}
& \mbox{0.657\,/\,0.660} \\

\textsc{Para}
& \mbox{0.007\,/\,0.050}
& \mbox{0.007\,/\,0.080}
& \mbox{0.063\,/\,0.213}
& \mbox{0.000\,/\,0.007}
& \mbox{0.413\,/\,0.607}
& \mbox{0.250\,/\,0.323} \\

\textsc{Combo}
& \mbox{0.953\,/\,0.570}
& \mbox{\textbf{1.000}\,/\,\textbf{1.000}}
& \mbox{\textbf{0.870}\,/\,\textbf{0.750}}
& \mbox{\textbf{0.747}\,/\,0.720}
& \mbox{\textbf{0.783}\,/\,\textbf{0.900}}
& \mbox{\textbf{0.757}\,/\,\textbf{0.993}} \\

\midrule
\rowcolor{gray!18}
\multicolumn{7}{c}{\textit{Qwen3-8B-Base}} \\
\midrule

\textsc{Act}
& \mbox{\textbf{1.000}\,/\,\textbf{1.000}}
& \mbox{\textbf{1.000}\,/\,\textbf{1.000}}
& \mbox{0.633\,/\,\textbf{0.633}}
& \mbox{\textbf{0.637}\,/\,0.637}
& \mbox{0.863\,/\,0.863}
& \mbox{0.540\,/\,0.603} \\

\textsc{Para}
& \mbox{0.000\,/\,0.020}
& \mbox{0.000\,/\,0.293}
& \mbox{0.020\,/\,0.110}
& \mbox{0.010\,/\,0.040}
& \mbox{0.083\,/\,0.370}
& \mbox{0.030\,/\,0.030} \\

\textsc{Combo}
& \mbox{0.527\,/\,0.847}
& \mbox{\textbf{1.000}\,/\,0.990}
& \mbox{\textbf{0.703}\,/\,0.543}
& \mbox{0.577\,/\,\textbf{0.767}}
& \mbox{\textbf{0.887}\,/\,\textbf{0.953}}
& \mbox{\textbf{0.603}\,/\,\textbf{0.887}} \\

\midrule
\rowcolor{gray!18}
\multicolumn{7}{c}{\textit{Llama-3.2-1B}} \\
\midrule

\textsc{Act}
& \mbox{\textbf{1.000}\,/\,\textbf{1.000}}
& \mbox{0.157\,/\,0.150}
& \mbox{0.017\,/\,0.020}
& \mbox{0.660\,/\,\textbf{0.667}}
& \mbox{0.413\,/\,0.607}
& \mbox{0.807\,/\,0.810} \\

\textsc{Para}
& \mbox{0.000\,/\,0.037}
& \mbox{0.003\,/\,0.693}
& \mbox{0.133\,/\,\textbf{0.230}}
& \mbox{0.057\,/\,0.263}
& \mbox{0.637\,/\,0.587}
& \mbox{0.467\,/\,0.457} \\

\textsc{Combo}
& \mbox{0.943\,/\,0.193}
& \mbox{\textbf{0.590}\,/\,\textbf{0.897}}
& \mbox{\textbf{0.317}\,/\,0.197}
& \mbox{\textbf{0.707}\,/\,0.633}
& \mbox{\textbf{0.973}\,/\,\textbf{0.783}}
& \mbox{\textbf{0.937}\,/\,\textbf{0.937}} \\

\bottomrule
\end{tabular*}
\endgroup

\caption{Exact-match accuracy across the six tasks, evaluated on 300 test instances per task with 50 history examples per instance. Each entry reports results with 2/20 adaptation steps for \textsc{Para} and \textsc{Combo}; \textsc{Act} is evaluated in two independent runs without parameter updates. Bold indicates the best result for each model, task, and adaptation budget, including ties.}
\label{tab:main_results}

\end{table*}




Across both update budgets, activation memory is stronger in factual recall, while parametric memory does not consistently outperform it in task learning, answering RQ1.

Table~\ref{tab:main_results} shows that \textsc{Act} achieves 1.000 Factual Recall accuracy across all four models, while \textsc{Para} remains near zero at both budgets. 
\textsc{Act} also consistently outperforms \textsc{Para} on Condition Judgment, while their relative advantage varies across other basic tasks, models, and budgets.
Increasing from 2 to 20 steps improves \textsc{Para} on several tasks but does not yield a consistent advantage. 
Extending adaptation to 200 steps on Factual Recall (Fig.~\ref{fig:memory_sensitivity}(a)) still gives only 4.67\% for \textsc{Para}, showing that its poor recall is not simply due to a limited update budget.

\subsection{Results for RQ2}

\begin{figure}[t]
    \centering
    \includegraphics[width=\columnwidth]
    {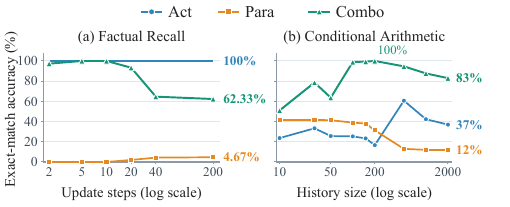}
    \caption{Qwen3-1.7B-Base accuracy on (a) Factual Recall across update steps with 50 history examples, and (b) Conditional Arithmetic across history sizes. 
    }
    \label{fig:memory_sensitivity}
    \vspace{-7pt}
\end{figure}

Conditional Arithmetic shows a consistent benefit from combining both memory types. 
Table~\ref{tab:main_results} shows that \textsc{Combo} achieves the highest accuracy across all four models at both update budgets, answering RQ2.
This advantage persists across history sizes.
On Qwen3-1.7B-Base, Fig.~\ref{fig:memory_sensitivity}(b) shows that \textsc{Combo} outperforms both individual memories from 10 to 2,000 history examples. 
It reaches 100\% accuracy with 200 examples and retains 83\% with 2,000, compared with 37\% for \textsc{Act} and 12\% for \textsc{Para}.

\subsection{Results for RQ3}

To investigate RQ3, we decompose Conditional Arithmetic into Condition Judgment and Arithmetic Execution. We first localize neurons associated with each component, then compare their activity across the three memory settings.

\subsubsection{Neuron Localization}
\label{sec:neuron_localization}

We use test instances from Conditional Arithmetic, Condition Judgment, and Arithmetic Execution defined in Section~\ref{sec:task_definitions}.
The analysis uses Qwen3-1.7B-Base with 50 history examples per instance and the 2-step setting for \textsc{Para} and \textsc{Combo}.

Comparing Conditional Arithmetic with Condition Judgment isolates the added execution component, while comparing it with Arithmetic Execution isolates the added condition component.
For each MLP neuron $u$ and test instance, we record its post-gating, pre-down-projection activation at the final prompt token before answer generation.
Let $\bar z_u^{(\mathrm{CA})}$, $\bar z_u^{(\mathrm{CJ})}$, and $\bar z_u^{(\mathrm{AE})}$ denote the mean activation of neuron $u$ across test instances for the three tasks, respectively. 
We define
\begin{equation}
S_{\mathrm{exec}}(u)=
\frac{\bar z_u^{(\mathrm{CA})}-\bar z_u^{(\mathrm{CJ})}}
{s_{\mathrm{CA},\mathrm{CJ}}(u)},\qquad
S_{\mathrm{cond}}(u)=
\frac{\bar z_u^{(\mathrm{CA})}-\bar z_u^{(\mathrm{AE})}}
{s_{\mathrm{CA},\mathrm{AE}}(u)}.
\label{eq:localization_scores}
\end{equation}

For each task $T\in\{\mathrm{CA},\mathrm{CJ},\mathrm{AE}\}$, let $\sigma_T^2(u)$ denote the sample variance of neuron $u$ across test instances. 
The pooled standard deviation is
$s_{\mathrm{CA},\mathrm{CJ}}(u)=\sqrt{(\sigma_{\mathrm{CA}}^2(u)+\sigma_{\mathrm{CJ}}^2(u))/2}$,
with $s_{\mathrm{CA},\mathrm{AE}}(u)$ defined analogously.
Higher scores indicate greater mean activation on Conditional Arithmetic relative to the comparison task (i.e., Condition Judgment or Arithmetic Execution), normalized by within-task variability.

Prior work reports functional differences across Qwen3
layers~\cite{chen2026propositional,deng2026qwenscope}.
We examine six prespecified layers of Qwen3-1.7B-Base.
Layers 15--18 form a contiguous middle block, with layer 2
as an early reference and layer 27 as a late reference.
Layer indices start at zero.

Within each layer, we select the 20 neurons with the highest score for each contrast, ensuring equal contribution from each analyzed layer. 
This yields 120 condition-associated and 120 execution-associated neurons.

\subsubsection{Neuronal Activity}
\label{sec:neuronal_activity}

We next compare the activity of the selected condition-associated and execution-associated neurons across \textsc{Act}, \textsc{Para}, and \textsc{Combo}. 
We use the same 300 Conditional Arithmetic test instances, each with 50 history examples, and follow the same activation measurement as in Section~\ref{sec:neuron_localization}: the post-gating, pre-down-projection activation at the final prompt token before answer generation.

To account for differences in activation scale across neurons, we normalize each neuron's activation by its sample standard deviation across the 300 test instances, computed separately for each memory setting. 
Let $z_{u,j}$ denote the activation of neuron $u$ on test instance $j$, and let $\sigma_u$ denote its standard deviation within the corresponding memory setting. 
We report
\begin{equation}
\mathrm{Mean}=\frac{1}{N}\sum_{u,j}\frac{z_{u,j}}{\sigma_u},\qquad
\mathrm{MeanAbs}=\frac{1}{N}\sum_{u,j}\frac{|z_{u,j}|}{\sigma_u}.
\label{eq:normalized_activity}
\end{equation}
For each neuron population and memory setting, the statistics average over $N=300\times120=36,000$ activation values. 
\textbf{Mean} preserves the sign of neuronal activity, whereas \textbf{MeanAbs} measures its normalized magnitude. 
Thus, these metrics characterize activity relative to each neuron's within-setting variability rather than its raw activation scale.


\begin{table}[!htbp]
\centering
\begingroup

\definecolor{MemModelBand}{HTML}{E6E6E6}
\definecolor{MemHigherActivity}{HTML}{E2F0D9}
\definecolor{MemLowerActivity}{HTML}{F8DADA}

\fontsize{9}{10.2}\selectfont
\setlength{\tabcolsep}{2.4pt}
\renewcommand{\arraystretch}{1.08}

\begin{tabularx}{\columnwidth}{
    @{}ll*{4}{>{\centering\arraybackslash}X}@{}
}
\toprule

\multirow{2}{*}{\textbf{Dataset}}
& \multirow{2}{*}{\textbf{Memory}}
& \multicolumn{2}{c}{\textbf{Condition}}
& \multicolumn{2}{c}{\textbf{Execution}} \\

\cmidrule(lr){3-4}
\cmidrule(lr){5-6}

& & Mean & MeanAbs & Mean & MeanAbs \\

\midrule
\rowcolor{white}
& \textsc{Act}
& \cellcolor{MemHigherActivity}1.112
& \cellcolor{MemHigherActivity}4.407
& \cellcolor{MemLowerActivity}0.413
& \cellcolor{MemLowerActivity}4.154 \\

\rowcolor{white}
& \textsc{Para}
& \cellcolor{MemLowerActivity}0.315
& \cellcolor{MemLowerActivity}3.071
& \cellcolor{MemHigherActivity}1.107
& \cellcolor{MemHigherActivity}5.782 \\

\rowcolor{white}
\multirow{-3}{*}{\shortstack{Conditional\\Arithmetic}}
& \textsc{Combo}
& \cellcolor{MemHigherActivity}1.183
& \cellcolor{MemHigherActivity}4.471
& \cellcolor{MemHigherActivity}0.923
& \cellcolor{MemHigherActivity}5.149 \\

\bottomrule
\end{tabularx}
\endgroup
\caption{Normalized activity of condition-associated and execution-associated neurons on Qwen3-1.7B-Base under the three memory settings. Green marks the two higher values and red the lowest value within each metric column.}
\label{tab:neuronal_activity}
\end{table}

Table~\ref{tab:neuronal_activity} reveals complementary activity patterns across the two neuron populations. 
For condition-associated neurons, \textsc{Act} and \textsc{Combo} show higher MeanAbs (4.407 and 4.471) than \textsc{Para} (3.071). 
In contrast, for execution-associated neurons, \textsc{Para} and \textsc{Combo} show higher MeanAbs (5.782 and 5.149) than \textsc{Act} (4.154).
Mean exhibits the same ordering in both populations.

Together, these results show that \textsc{Combo} exhibits the stronger condition-associated activity observed with \textsc{Act} and the stronger execution-associated activity observed with \textsc{Para}.
This complementary pattern accompanies the higher Conditional Arithmetic accuracy of \textsc{Combo} (63.67\%), compared with 25.67\% for \textsc{Act} and 41.33\% for \textsc{Para}. 
The neuronal activity patterns therefore align with the two components of Conditional Arithmetic and provide neuron-level insight into how the two memory types complement each other, answering RQ3.



\section{Conclusion}
\label{sec:conclusion}


We systematically study the roles of activation and parametric memory in test-time learning. 
Activation memory is consistently effective for factual recall, while parametric memory does not consistently outperform it in task learning, challenging a simple division of roles between the two memory types. 
For the composite task, however, combining both memories yields substantial gains. 
Neuron-level analysis further reveals complementary activity patterns associated with condition judgment and task execution.
Together, these results shed light on how activation and parametric memory can complement each other, providing a step toward understanding the mechanisms underlying test-time learning.

\bibliographystyle{IEEEbib}
\bibliography{strings,refs}


\section*{\centering\normalsize\bfseries COMPLIANCE WITH ETHICAL STANDARDS}
This study uses synthetic data and does not involve human
participants or animal subjects. Ethical approval was not required.

\end{document}